\documentclass[runningheads]{llncs}
\usepackage{graphicx}

\usepackage{subfig}
\usepackage{amssymb}
\usepackage{comment}
\usepackage{amsmath}
\usepackage{multirow}
\usepackage{soul}

\usepackage{amsmath}

\begin{document}
%
\title{Gated Convolutional Neural Networks \\ for Domain Adaptation}
%
%

\author{
Avinash Madasu and Vijjini Anvesh Rao}
\institute{Samsung R\&D Institute, Bangalore\\
           \email{\{m.avinash,a.vijjini\}@samsung.com}}

\maketitle

\begin{abstract}
Domain Adaptation explores the idea of how to maximize performance on a target domain, distinct from source domain, upon which the classifier was trained. This idea has been explored for the task of sentiment analysis extensively. The training of reviews pertaining to one domain and evaluation on another domain is widely studied for modeling a domain independent algorithm. This further helps in understanding co-relation between domains. In this paper, we show that Gated Convolutional Neural Networks (GCN) perform effectively at learning sentiment analysis in a manner where domain dependant knowledge is filtered out using its gates. We perform our experiments on multiple gate architectures: Gated Tanh ReLU Unit (GTRU), Gated Tanh Unit (GTU) and Gated Linear Unit (GLU). Extensive experimentation on two standard datasets relevant to the task, reveal that training with Gated Convolutional Neural Networks give significantly better performance on target domains than regular convolution and recurrent based architectures. While complex architectures like attention, filter domain specific knowledge as well, their complexity order is remarkably high as compared to gated architectures. GCNs rely on convolution hence gaining an upper hand through parallelization.

\end{abstract}

\section{Introduction}
With the advancement in technology and invention of modern web applications like Facebook and Twitter, users started expressing their opinions and ideologies at a scale unseen before. The growth of e-commerce companies like Amazon, Walmart have created a revolutionary impact in the field of consumer business. People buy products online through these companies and write reviews for their products. These consumer reviews act as a bridge between consumers and companies. Through these reviews, companies polish the quality of their services. Sentiment Classification (SC) is one of the major applications of Natural Language Processing (NLP) which aims to find the polarity of text. In the early stages \cite{pang2002thumbs} of text classification, sentiment classification was performed using traditional feature selection techniques like Bag-of-Words (BoW) \cite{harris1954distributional} or TF-IDF. These features were further used to train machine learning classifiers like Naive Bayes (NB) \cite{mccallum1998comparison} and Support Vector Machines (SVM)\cite{joachims1998text}. They are shown to act as strong baselines for text classification \cite{wang2012baselines}. However, these models ignore  word level semantic knowledge and sequential nature of text. Neural networks were proposed to learn distributed representations of words \cite{bengio2003neural}. Skip-gram and CBOW architectures \cite{mikolov2013distributed} were introduced to learn high quality word representations which constituted a major breakthrough in NLP. Several neural network architectures like recursive neural networks \cite{socher2011parsing} and convolutional neural networks \cite{kim2014convolutional} achieved excellent results in text classification. Recurrent neural networks which were proposed for dealing sequential inputs suffer from vanishing \cite{bengio1994learning} and exploding gradient problems \cite{pascanu2013difficulty}. To overcome this problem, Long Short Term Memory (LSTM) was introduced \cite{hochreiter1997long}. 

All these architectures have been successful in performing sentiment classification for a specific domain utilizing large amounts of labelled data. However, there exists insufficient labelled data for a target domain of interest. Therefore, Domain Adaptation (DA) exploits knowledge from a relevant domain with abundant labeled data to perform sentiment classification on an unseen target domain. However, expressions of sentiment vary in each domain. For example, in $\textit{Books}$ domain, words $\textit{thoughtful}$ and $\textit{comprehensive}$ are used to express sentiment whereas $\textit{cheap}$ and $\textit{costly}$ are used in $\textit{Electronics}$ domain. Hence, models should generalize well for all domains. Several methods have been introduced for performing Domain Adaptation. Blitzer \cite{blitzer2006domain} proposed Structural Correspondence Learning (SCL) which relies on pivot features between source and target domains. Pan \cite{pan2010cross} performed Domain Adaptation using Spectral Feature Alignment (SFA) that aligns features across different domains. Glorot \cite{glorot2011domain} proposed Stacked Denoising Autoencoder (SDA) that learns generalized feature representations across domains. Zheng \cite{li2017end} proposed end-to-end adversarial network for Domain Adaptation. Qi \cite{liu2018learning} proposed a memory network for Domain Adaptation. Zheng \cite{li2018hierarchical} proposed a Hierarchical transfer network relying on attention for Domain Adaptation.

However, all the above architectures use a different sub-network altogether to incorporate domain agnostic knowledge and is combined with main network in the final layers. This makes these architectures computationally intensive. To address this issue, we propose a Gated Convolutional Neural Network (GCN) model that learns domain agnostic knowledge using gated mechanism \cite{dauphin2017language}. Convolution layers learns the higher level representations for source domain and gated layer selects domain agnostic representations. Unlike other models, GCN doesn't rely on a special sub-network for learning domain agnostic representations. As, gated mechanism is applied on Convolution layers, GCN is computationally efficient.
\begin{figure}
\centering
{\includegraphics[width=6cm]{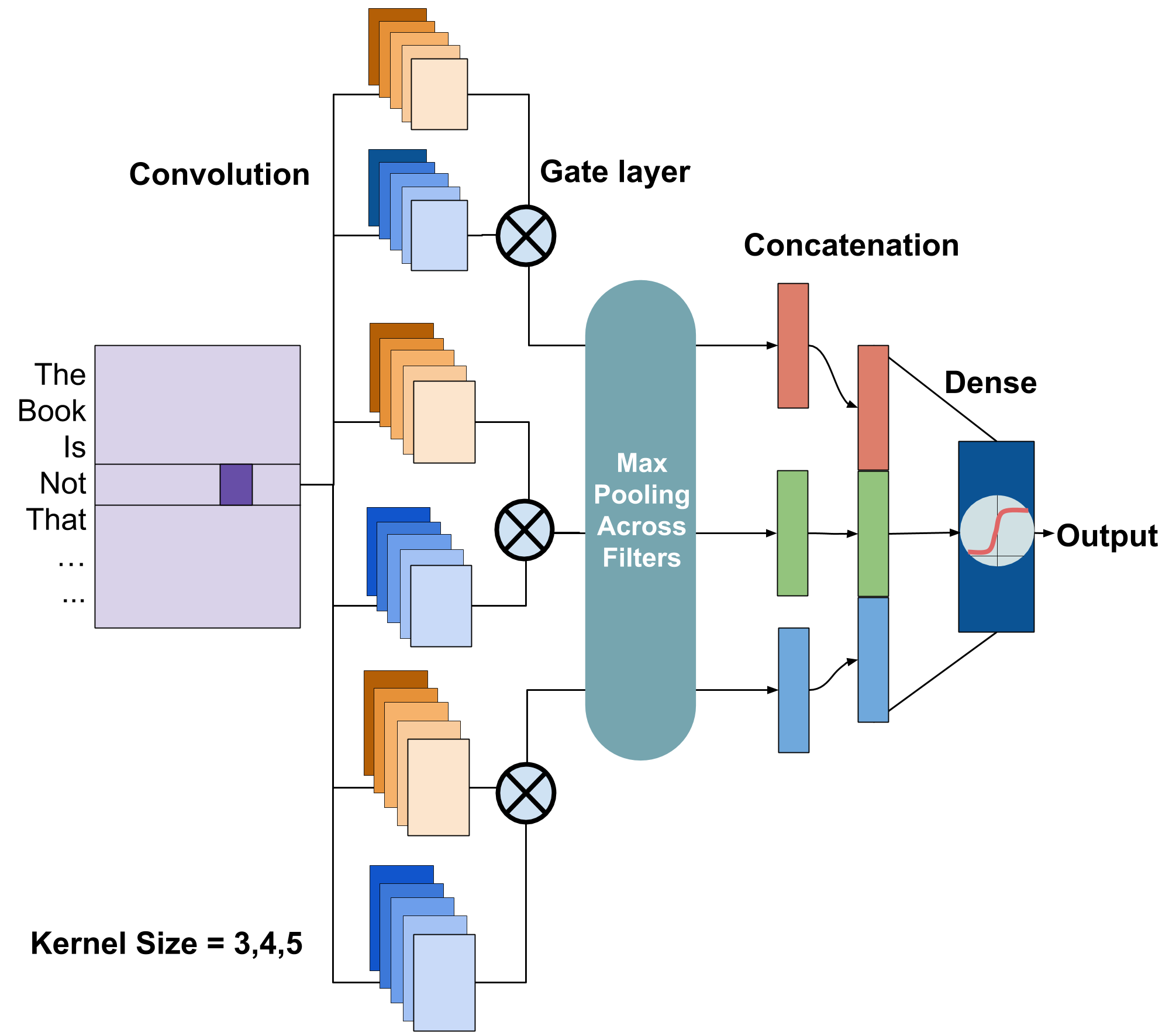}}
\caption{Architecture of the proposed model}
\label{fig:arch}
\end{figure}
\section{Related Work}
Traditionally methods for tackling Domain Adaptation are lexicon based. Blitzer \cite{blitzer2007biographies} used a pivot method to select features that occur frequently in both domains. It assumes that the selected pivot features can reliably represent the source domain. The pivots are selected using mutual information between selected features and the source domain labels. SFA \cite{pan2010cross} method argues that pivot features selected from source domain cannot attest a representation of target domain. Hence, SFA tries to exploit the relationship between domain-specific and domain independent words via simultaneously co-clustering them in a common latent space. SDA \cite{glorot2011domain} performs Domain Adaptation by learning intermediate representations through auto-encoders. Yu \cite{yu2016learning} used two auxiliary tasks to help induce sentence embeddings that work well across different domains. These embeddings are trained using Convolutional Neural Networks (CNN). 

Gated convolutional neural networks have achieved state-of-art results in language modelling \cite{dauphin2017language}. Since then, they have been used in different areas of natural language processing (NLP) like sentence similarity \cite{Chen2018GatedCN} and aspect based sentiment analysis \cite{xue2018aspect}.

\section{Gated Convolutional Neural Networks}
In this section, we introduce a model based on Gated Convolutional Neural Networks for Domain Adaptation. We present the problem definition of Domain Adaptation, followed by the architecture of the proposed model.

\subsection{Problem Definition}
Given a source domain $D_{S}$ represented as $D_{S}$ = \{$(x_{s_{1}},y_{s_{1}})$,$(x_{s_{2}},y_{s_{2}})$....$(x_{s_{n}},y_{s_{n}})$\} where  $x_{s_{i}} \in \mathbb{R}$ represents the vector of $i^{th}$ source text and $y_{s_{i}}$ represents the corresponding source domain label. Let $T_{S}$ represent the task  in source domain. Given a target domain $D_{T}$ represented as $D_{T}$ = \{$(x_{t_{1}},y_{t_{1}})$,$(x_{t_{2}},y_{t_{2}})$....$(x_{t_{n}},y_{t_{n}})$\}, where  $x_{t_{i}} \in \mathbb{R}$ represents the vector of $i^{th}$ target text and $y_{t_{i}}$ represents corresponding target domain label. Let $T_{T}$ represent the task in target domain. Domain Adaptation (DA) is defined by the target predictive function $f_{T}(D_{T})$ calculated using the knowledge of $D_{S}$ and $T_{S}$ where $D_{S} \neq D_{T}$ but $T_{S} = T_{T}$. It is imperative to note that the domains are different but only a single task. In this paper, the task is sentiment classification.
\subsection{Model Architecture}
The proposed model architecture is shown in the Figure \ref{fig:arch}.
Recurrent Neural Networks like LSTM, GRU update their weights at every timestep sequentially and hence lack parallelization over inputs in training. In case of attention based models, the attention layer has to wait for outputs from all timesteps. Hence, these models fail to take the advantage of parallelism either. Since, proposed model is based on convolution layers and gated mechanism, it can be parallelized efficiently. The convolution layers learn higher level representations for the source domain. The gated mechanism learn the domain agnostic representations. They together control the information that has to flow through further fully connected output layer after max pooling. 

Let $I$ denote the input sentence represented as $I$ = \{$w_{1}$$w_{2}$$w_{3}$...$w_{N}$\} where $w_{i}$ represents the $i_{th}$ word in $I$ and $N$ is the maximum sentence length considered. Let $B$ be the vocabulary size for each dataset and $X \in \mathbb{R}^{B \times d}$ denote the word embedding matrix where each $X_{i}$ is a $d$ dimensional vector. Input sentences whose length is less than $N$ are padded with 0s to reach maximum sentence length. Words absent in the pretrained word embeddings\footnote{https://nlp.stanford.edu/data/glove.840B.300d.zip} are initialized to 0s. Therefore each input sentence $I$ is converted to $P \in \mathbb{R}^{N \times d}$ dimensional vector. Convolution operation is applied on $P$ with kernel $K \in \mathbb{R}^{h \times d}$. The convolution operation is one-dimensional, applied with a fixed window size across words. We consider kernel size of 3,4 and 5. The weight initialization of these kernels is done using glorot uniform \cite{glorot2010understanding}. Each kernel is a feature detector which extracts patterns from n-grams. After convolution we obtain a new feature map $C$ = [$c_{1}c_{2}..c_{N}$] for each kernel $K$.
\begin{equation}
    C_{i} = f(P_{i:i+h} \ast W_{a} + b_{a})
\end{equation}
where $f$ represents the activation function in convolution layer.
The gated mechanism is applied on each convolution layer. Each gated layer learns to filter domain agnostic representations for every time step $i$.
\begin{equation}
    S_{i} = g(P_{i:i+h} \ast W_{s} + b_{s})
\end{equation}
where $g$ is the activation function used in gated convolution layer.
The outputs from convolution layer and gated convolution layer are element wise multiplied to compute a new feature representation $G_{i}$
\begin{equation}
    G_{i} = C_{i} \times S_{i}
\end{equation}

\begin{figure}%
    \centering
    \subfloat[GTRU]{{\includegraphics[width=3cm]{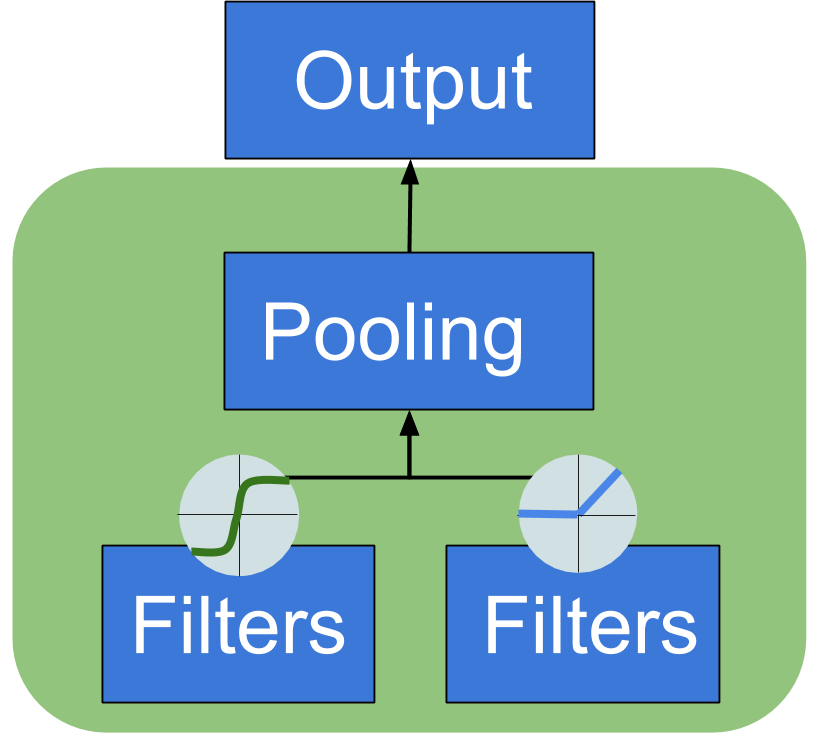} }}%
    \subfloat[GTU]{{\includegraphics[width=3cm]{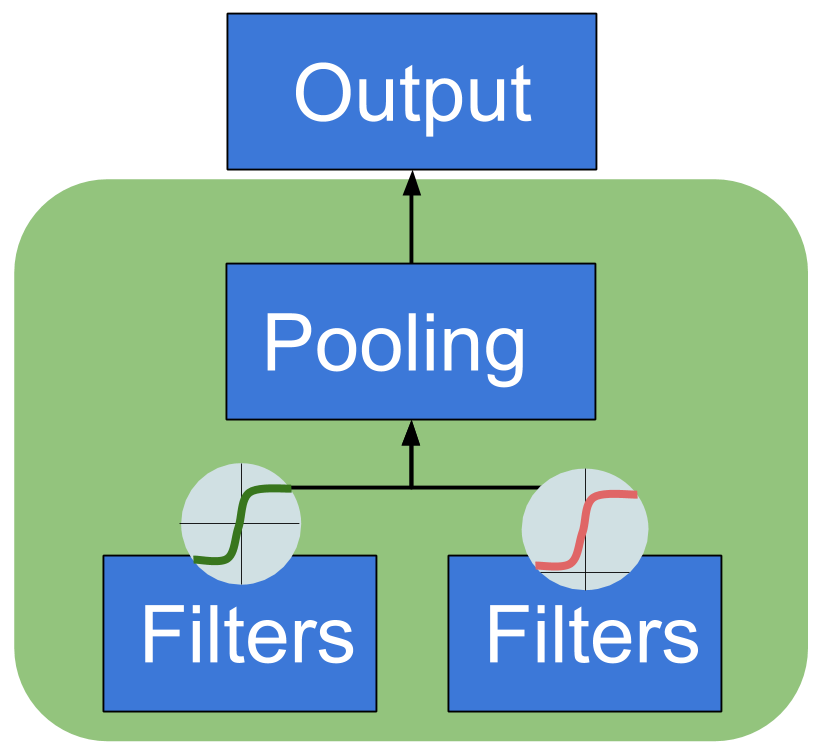} }}%
    \subfloat[GLU]{{\includegraphics[width=3cm]{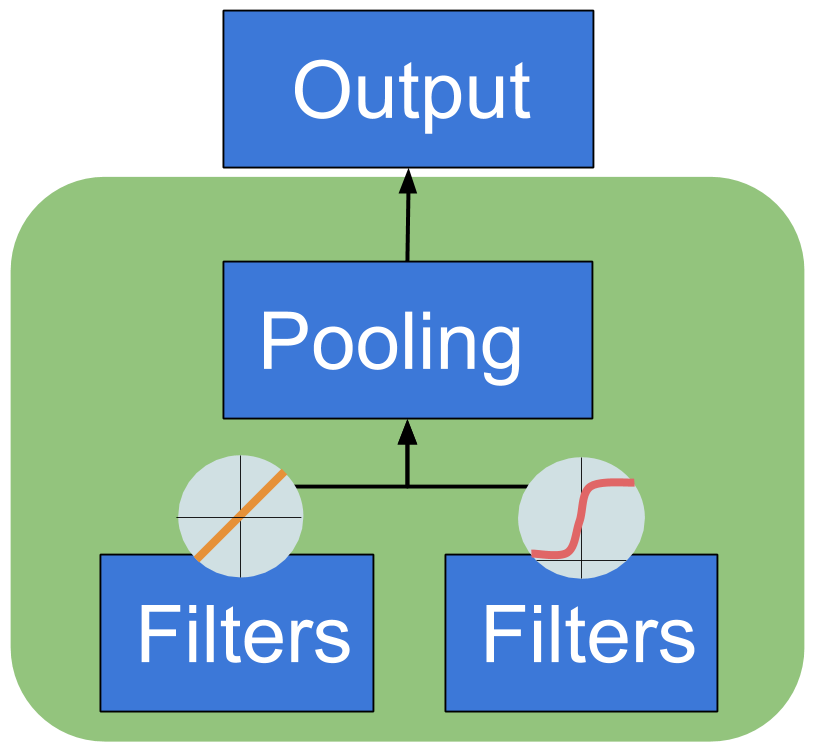} }}%
    \caption{Variations in gates of the proposed GCN architecture.}
    \label{fig:GU}
\end{figure}
Maxpooling operation is applied across each filter in this new feature representation to get the most important features \cite{kim2014convolutional}. As shown in Figure \ref{fig:arch} the outputs from maxpooling layer across all filters are concatenated. The concatenated layer is fully connected to output layer. Sigmoid is used as the activation function in the output layer. 

\subsection{Gating mechanisms}
Gating mechanisms have been effective in Recurrent Neural Networks like GRU and LSTM. They control the information flow through their recurrent cells. In case of GCN, these gated units control the domain information that flows to pooling layers. The model must be robust to change in domain knowledge and should be able to generalize well across different domains. We use the gated mechanisms Gated Tanh Unit (GTU) and Gated Linear Unit (GLU) and Gated Tanh ReLU Unit (GTRU) \cite{xue2018aspect} in proposed model. The gated architectures are shown in figure \ref{fig:GU}.
The outputs from Gated Tanh Unit is calculated as $tanh(P \ast W + c) \times \sigma(P \ast V + c)$. In case of Gated Linear Unit, it is calculated as $(P \ast W + c) \times \sigma(P \ast V + c)$ where $tanh$ and $\sigma$ denotes Tanh and Sigmoid activation functions respectively. In case of Gated Tanh ReLU Unit, output is calculated as  $tanh(P \ast W + c) \times relu(P \ast V + c)$
\section{Experiments}
\subsection{Datasets}
\subsubsection{Multi Domain Dataset(MDD)}
Multi Domain Dataset \cite{blitzer2007biographies} is a short dataset with reviews from distinct domains namely Books(B), DVD(D), Electronics(E) and Kitchen(K). Each domain consists of 2000 reviews equally divided among positive and negative sentiment. We consider 1280 reviews for training, 320 reviews for validation and 400 reviews for testing from each domain.
\subsubsection{Amazon Reviews Dataset(ARD)}
Amazon Reviews Dataset \cite{he2016ups} is a large dataset with millions of reviews from different product categories. For our experiments, we consider a subset of 20000 reviews from the domains Cell Phones and Accessories(C), Clothing and Shoes(S), Home and Kitchen(H) and Tools and Home Improvement(T). Out of 20000 reviews, 10000 are positive and 10000 are negative. We use 12800 reviews for training, 3200 reviews for validation and 4000 reviews for testing from each domain.
\subsection{Baselines}
To evaluate the performance of proposed model, we consider various baselines like traditional lexicon approaches, CNN models without gating mechanisms and LSTM models.
\subsubsection{BoW+LR}
Bag-of-words (BoW) is one of the strongest baselines in text classification \cite{wang2012baselines}. We consider all the words as features with a minimum frequency of 5. These features are trained using Logistic Regression (LR). 
\subsubsection{TF-IDF+LR}
TF-IDF is a feature selection technique built upon Bag-of-Words. We consider all the words with a minimum frequency of 5. The features selected are trained using Logistic Regression (LR).
\subsubsection{PV+FNN}
Paragraph2vec or doc2vec \cite{le2014} is a strong and popularly used baseline for text classification. Paragraph2Vec represents each sentence or paragraph in the form of a distributed representation. We trained our own doc2vec model using DBOW model. The paragraph vectors obtained are trained using Feed Forward Neural Network (FNN).  
\subsubsection{CNN}
To show the effectiveness of gated layer, we consider a CNN model which does not contain gated layers. Hence, we consider Static CNN model, a popular CNN architecture proposed in Kim \cite{kim2014convolutional} as a baseline.
\subsubsection{CRNN}
Wang \cite{wang2016combination} proposed a combination of Convolutional and Recurrent Neural Network for sentiment Analysis of short texts. This model takes the advantages of features learned by CNN and long-distance dependencies learned by RNN. It achieved remarkable results on benchmark datasets. We report the results using code published by the authors\footnote{ https://github.com/ultimate010/crnn}.
\subsubsection{LSTM}
We offer a comparison with LSTM model with a single hidden layer. This model is trained with equivalent experimental settings as proposed model.
\subsubsection{LSTM+Attention}
In this baseline, attention mechanism \cite{bahdanau2014neural} is applied on the top of LSTM outputs across different timesteps.
\begin{table}
\centering
\setlength{\tabcolsep}{12pt}
\caption{Average training time for all the models on ARD}
\label{tab:time}
\begin{tabular}{|c|c|c|}
\hline
    Model & Batchsize & Time for 1 epoch(in Sec)  \\
    \hline
    CRNN & 50 & 50  \\ 
     \hline
     LSTM & 50& 70 \\ 
     \hline
     LSTM+Attention&50 & 150 \\ 
     \hline
    GLU & 50& 10 \\ 
     \hline
     GRU&50 & 10 \\ 
     \hline
     GTRU&50 & 10 \\ 
     \hline
\end{tabular}
\end{table}

\subsection{Implementation details}
All the models are experimented with approximately matching number of parameters for a solid comparison using a Tesla K80 GPU.\\ \\
{\bf Input}
Each word in the input sentence is converted to a 300 dimensional vector using GloVe pretrained vectors \cite{pennington2014glove}. A maximum sentence length 100 is considered for all the datasets. Sentences with length less than 100 are padded with 0s.\\ \\
{\bf Architecture details}:
The model is implemented using keras. We considered 100 convolution filters for each of the kernels of sizes 3,4 and 5. To get the same sentence length after convolution operation zero padding is done on the input. \\ \\
{\bf Training}
Each sentence or paragraph is converted to lower case. Stopword removal is not done. A vocabulary size of 20000 is considered for all the datasets. We apply a dropout layer \cite{srivastava2014dropout} with a probability of 0.5, on the embedding layer and probability 0.2, on the dense layer that connects the output layer. Adadelta \cite{zeiler2012adadelta} is used as the optimizer for training with gradient descent updates. Batch-size of 16 is taken for MDD and 50 for ARD. The model is trained for 50 epochs. We employ an early stopping mechanism based on validation loss for a patience of 10 epochs. The models are trained on source domain and tested on unseen target domain in all experiments.\begin{table}
\centering
\setlength{\tabcolsep}{9pt}
\caption{Accuracy scores on Multi Domain Dataset.}
\label{tab:multidomain1}
\begin{tabular}{|c|c|c|c|c|c|c|}
\hline
Source$->$Target & BoW & TFIDF & PV & CNN & CRNN & LSTM  \\
\hline
B$->$D&72.5 &73.75 & 63.749&57.75 & 68.75&69.5 \\ 
\hline
B$->$E&67.5 &68.5 &53.25 & 53.5& 63.249&58.75 \\ 
\hline
B$->$K&69.25 &72.5 &57.75 &56.25 & 66.5&64.75 \\ 
\hline
D$->$B&66 &68.5 & 64.75&54.25 &66.75&74.75 \\ 
\hline
D$->$E&71 &69.5 &56.75 &57.25 &69.25 &64.25 \\ 
\hline
D$->$K&68 &69.75 &60 & 58.25&67.5 &70 \\ 
\hline
E$->$B&63.249 &64 &54 &57.25 &69.5 &67.75 \\ 
\hline
E$->$D&65 &66 &47.25 &56.499 &64.5 & 67\\ 
\hline
E$->$K&76.25 &76.75 &59.25 &63.249 & 76&76 \\ 
\hline
K$->$B&61.5 &67.75 &50 &57.75 & 69.25&66.25 \\ 
\hline
K$->$D&68 &70.5 & 52.25&60 &64.75 &71 \\ 
\hline
K$->$E&81 &80 &50 &59.25 & 69& 76.75\\ 
\hline
\end{tabular}
\end{table}

\begin{table}
\centering
\setlength{\tabcolsep}{11pt}
\caption{Accuracy scores on Multi Domain Dataset.}
\label{tab:multidomain2}
\begin{tabular}{|c|c|c|c|c|}
\hline
Source$->$Target & LSTM.Attention & GLU & GTU & GTRU  \\
\hline
B$->$D&76.75 &\textbf{79.5} &79.25 & 77.5  \\ 
\hline
B$->$E&70 &\textbf{71.75} &71.25 & 71.25  \\ 
\hline
B$->$K&74.75 & 73&72.5 & \textbf{74.25} \\ 
\hline
D$->$B&72.5 &78 &\textbf{80.25} & 77.25  \\ 
\hline
D$->$E&71 &73 &\textbf{74.5} & 69.25  \\ 
\hline
D$->$K&72.75 &\textbf{77} &76 & 74.75  \\ 
\hline
E$->$B& 64.75&\textbf{71.75} &68.75 & 67.25  \\ 
\hline
E$->$D&62.749 &\textbf{71.75} &69 & 68.25  \\ 
\hline
E$->$K&72 &\textbf{82.25} & 80.5& 79  \\ 
\hline
K$->$B&64.75 &\textbf{70} &67.75 & 63.249  \\ 
\hline
K$->$D&\textbf{75} &73.75 & 73.5& 69.25  \\ 
\hline
K$->$E& 75.5&\textbf{82} &\textbf{82} & 81.25  \\ 
\hline
\end{tabular}
\end{table}

\begin{table}
\centering
\setlength{\tabcolsep}{9pt}
\caption{Accuracy scores on Amazon Reviews Dataset.}
\label{tab:amazon1}
\begin{tabular}{|c|c|c|c|c|c|c|}
\hline
Source$->$Target & BoW & TFIDF & PV & CNN & CRNN & LSTM  \\
\hline
C$->$S & 79.3&81.175 &69.625 &62.324 &84.95 &83.7 \\
\hline
C$->$H & 81.6&82.875 & 70.775& 59.35&81.8 & 81.175\\
\hline
C$->$T &76.25 &77.475 & 66.4& 54.5& 79.025&77.175 \\
\hline
S$->$C &76.925 & 76.525&69.425 & 55.375& 79.975&79.85 \\
\hline
S$->$T & 80.125&81.575 &74.524 &62.7 &81.45 & 82.925\\
\hline
S$->$H &74.275 &75.175 &67.274 &61.925 &76.05 &77.7 \\
\hline
H$->$S &76.149 & 73.575&65.3 &53.55 &79.074 & 78.574\\
\hline
H$->$C &81.225 & 80.925&70.7 & 58.25&74.275 &81.95 \\
\hline
H$->$T &79.175 &75.449 &69.425 &59.4 & 76.325&76.725 \\
\hline
T$->$C &75.1 &73.875 &56.85 &56 &80.25 &76.9 \\
\hline
T$->$S &78.875 &80.5 & 59.199&60 &\textbf{85.824} & 81.8\\
\hline
T$->$H &81.325 &81.875 &66.8 &61.25 & 83.35& 81\\
\hline
\end{tabular}
\end{table}
\begin{table}
\centering
\setlength{\tabcolsep}{9pt}
\caption{Accuracy scores on Amazon Reviews Dataset.}
\label{tab:amazon2}
\begin{tabular}{|c|c|c|c|c|c|c|}
\hline
Source$->$Target & LSTM.Attention & GLU & GTU & GTRU \\
\hline
C$->$S &84.15 &\textbf{85.125} & 84.95&84.8 \\
\hline
C$->$H & 82.6& \textbf{84.85}& 84.2& 84.55\\
\hline
C$->$T & 77.9& 79.5& 79.274&\textbf{80.225} \\
\hline
S$->$C & 78.075&80.925 &80.25 & \textbf{83.1}\\
\hline
S$->$H & 82.325&83.95 &83.399 & \textbf{84.025}\\
\hline
S$->$T &78.425 & \textbf{79.475}&77.85 & 79.375\\
\hline
H$->$C & 81.375&\textbf{83.175} & 81.85& 82.1\\
\hline
H$->$S & 81.975&82.75 & 84.1& \textbf{85.425}\\
\hline
H$->$T &80.95 &\textbf{82.55} & 81.774&81.825 \\
\hline
T$->$C &75.55 & \textbf{82.125}&80.805 &81.825 \\
\hline
T$->$S & 82.375&82.625 &83.975 &84.775 \\
\hline
T$->$H & 80.5&84.7 &83.95 & \textbf{85.275}\\
\hline

\end{tabular}
\end{table}

\section{Results and Discussion}

\subsection{Results}
The performance of all models on MDD is shown in Tables \ref{tab:multidomain1} and \ref{tab:multidomain2} while for ARD, in Tables \ref{tab:amazon1} and \ref{tab:amazon2}. All values are shown in accuracy percentage. Furthermore time complexity of each model is presented in Table \ref{tab:time}.  
\subsection{Discussion}
\subsubsection{Gated outperform regular Convolution}
We find that gated architectures vastly outperform non gated CNN model. The effectiveness of gated architectures rely on the idea of training a gate with sole purpose of identifying a weightage. In the task of sentiment analysis this weightage corresponds to what weights will lead to a decrement in final loss or in other words, most accurate prediction of sentiment. In doing so, the gate architecture learns which words or n-grams contribute to the sentiment the most, these words or n-grams often co-relate with domain independent words. On the other hand the gate gives less weightage to n-grams which are largely either specific to domain or function word chunks which contribute negligible to the overall sentiment. This is what makes gated architectures effective at Domain Adaptation.

\begin{figure}%
    \centering
    \subfloat[``good cell phone'']{{\includegraphics[width=5.5cm]{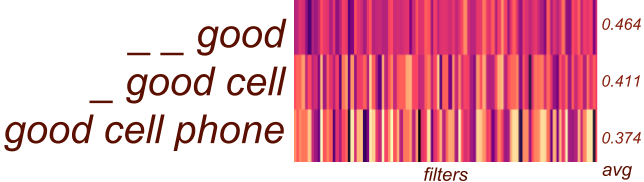} }}%
    \hfill
    \subfloat[``costly would have to sell entire kitchen'']{{\includegraphics[width=6cm]{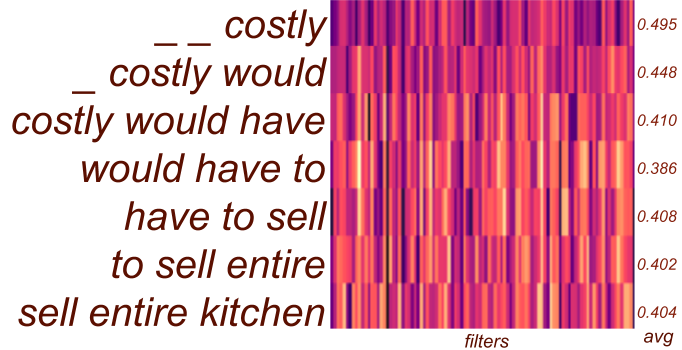} }}%
    \qquad
    \hfill
    \subfloat[``great quality functional case'']{{\includegraphics[width=6cm]{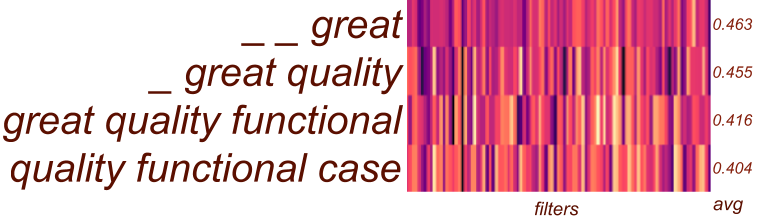} }}%
    \caption{Visualizing outputs from gated convolutions (filter size = 3) of GLU for example sentences, darker indicates higher weightage}%
    \label{fig:example}%
\end{figure}

In Figure \ref{fig:example}, we have illustrated the visualization of convolution outputs(kernel size = 3) from the sigmoid gate in GLU across domains. As the kernel size is 3, each row in the output corresponds to a trigram from input sentence. This heat map visualizes values of all 100 filters and their average for every input trigram. These examples demonstrate what the convolution gate learns. Trigrams with domain independent but heavy polarity like ``\textunderscore \hspace{0.05cm} \textunderscore \hspace{0.05cm} good'' and ``\textunderscore \hspace{0.05cm} costly would'' have higher weightage. 
Meanwhile, Trigrams with domain specific terms like ``quality functional case'' and ``sell entire kitchen'' get some of the least weights. 
In Figure \ref{fig:example}(b) example, the trigram ``would have to'' just consists of function words, hence gets the least weight. While ``sell entire kitchen'' gets more weight comparatively. This might be because while function words are merely grammatical units which contribute minimal to overall sentiment, domain specific terms like ``sell'' may contain sentiment level knowledge only relevant within the domain. In such a case it is possible that the filters effectively propagate sentiment level knowledge from domain specific terms as well.
\subsubsection{Gated outperform Attention and Linear} 
We see that gated architectures almost always outperform recurrent, attention and linear models BoW, TFIDF, PV. This is largely because while training and testing on same domains, these models especially recurrent and attention based may perform better. However, for Domain Adaptation, as they lack gated structure which is trained in parallel to learn importance, their performance on target domain is poor as compared to gated architectures. As gated architectures are based on convolutions, they exploit parallelization to give significant boost in time complexity as compared to other models. This is depicted in Table \ref{tab:time}.
\subsubsection{Comparison among gates}
While the gated architectures outperform other baselines, within them as well we make observations. Gated Linear Unit (GLU) performs the best often over other gated architectures. In case of GTU, outputs from Sigmoid and Tanh  are multiplied together, this may result in small gradients, and hence resulting in the vanishing gradient problem. However, this will not be the in the case of GLU, as the activation is linear. In case of GTRU, outputs from Tanh and ReLU are multiplied. In ReLU, because of absence of negative activations, corresponding Tanh outputs will be completely ignored, resulting in loss of some domain independent knowledge. 
\section{Conclusion}
In this paper, we proposed Gated Convolutional Neural Network(GCN) model for Domain Adaptation in Sentiment Analysis. We show that gates in GCN, filter out domain dependant knowledge, hence performing better at an unseen target domain. Our experiments reveal that gated architectures outperform other popular recurrent and non-gated architectures. Furthermore, because these architectures rely on convolutions, they take advantage of parellalization, vastly reducing time complexity.

\bibliographystyle{splncs04}
\bibliography{llncs}

\end{document}